\documentclass[runningheads]{llncs}
\usepackage{graphicx}
\graphicspath{images}
\usepackage{cite}
\usepackage{amsmath,amssymb,amsfonts}
\usepackage{algorithmic}
\usepackage{graphicx}
\usepackage{textcomp}
\usepackage{xcolor}
\usepackage[tracking=true]{microtype}
\usepackage{array}
\usepackage{multirow}
\usepackage{enumitem}

\newcommand{\smallvsqueaze}{\vspace{-1.5mm}}
\newcommand{\vsqueaze}{\vspace{-3mm}}
\newcommand{\bigvsqueaze}{\vspace{-5mm}}

\begin{document}
\title{Playing Catan with Cross-dimensional Neural Network}
%
%
\author{Quentin Gendre\inst{1}\orcidID{0000-0003-3352-4321} \and
Tomoyuki Kaneko\inst{2}\orcidID{0000-0001-8051-2388}}
\authorrunning{Q. Gendre and T. Kaneko}
%
\institute{Graduate School of Interdisciplinary Information Studies, The University of Tokyo, Japan
\email{gendre@g.ecc.u-tokyo.ac.jp}\\
\and
Interfaculty Initiative in Information Studies, The University of Tokyo, Japan
\email{kaneko@graco.c.u-tokyo.ac.jp}\\
\url{http://www.graco.c.u-tokyo.ac.jp/\homedir kaneko/}}
\maketitle              
\begin{abstract}
Catan\footnote{previously named The Settlers of Catan, renamed for the 5th Edition (2015)} is a strategic board game with many interesting properties, including multi-player, imperfect information, stochasticity, a complex state space structure (hexagonal board where each vertex, edge and face has its own features, cards for each player, etc), and a large action space (including trading). %
Therefore, it is challenging to build AI agents by Reinforcement Learning (RL), without domain knowledge nor heuristics.
In this paper, we introduce cross-dimensional neural networks to handle a mixture of information sources and a wide variety of outputs, and empirically demonstrate that the network dramatically improves RL in Catan.
We also show that, for the first time, a RL agent can outperform \textit{jsettler}, the best heuristic agent available.

\end{abstract}

\section{Introduction}

Among the challenges toward practical AI agents in real world, this paper focuses on three:
\begin{itemize}
\item learning a task with a general method and no prior domain-specific knowledge
\item handling information sources of different kinds (e.g. not only images)
\item acting robustly even when only a part of the world can be observed
\end{itemize}
Games have long served as testbeds for AI research, and recently 
AlphaZero~\cite{AlphaZero} presented a general reinforcement learning method that successfully mastered chess, shogi, and Go, without human knowledge. 
However, these are deterministic perfect information games,
where agents can utilize Monte-Carlo tree search (MCTS), and they have simple representations, a square grid board easily handled by standard image recognition techniques (e.g. CNN). Therefore, it is not clear whether that method would work well in more complex domains with imperfect information, non-deterministic, and/or complex state representation, where neither MCTS nor CNN are directly available. 
In this paper, we study Catan\footnotemark[\value{footnote}] -- a famous Euro-style board game that has sold more than 22 million copies, and with frequent international tournaments -- as a representative of such complex domains. Catan is an imperfect information and non-deterministic game, in which agents need to handle multiple observations (hexagonal board, cards for each player, etc...) as well as guessing hidden information and possible futures depending on opponents' resources and randomness.
We integrated a standard policy gradient method in deep reinforcement learning with self-play, and introduced cross-dimensional network, a network structure that supports multiple input sources in a flexible manner, that empirically outperformed the baseline \emph{jsettler}.




\section{Background and Related Work}

\subsection{Deep Reinforcement Learning in Two-Player Games}
We follow standard notation of reinforcement learning; where an agent learns via interaction with an environment.  For details, readers are referred to a textbook~\cite{SuttonBarto2018}.
Usually, an environment is modeled as Markov Decision Process (MDP), $(S,A,T,R,\gamma)$, though many applications of RL are not conforming Markov property in practice. 
At each time step $t$, an agent observes a state $s_t \in S$,
and chooses an action $a\in A$.  The environment changes its state to $s_{t+1}$ following transition function $T$, and the agent receives a reward $r_t$. 
The policy $\pi: S \times A \mapsto \mathbb{R}$ of an agent is a probability distribution over actions given an observation.
The (ultimate) goal of the learning is to identify the optimal policy $\pi^*$ that maximizes the expected cumulative rewards $\mathbb{E}_{a\sim \pi^*}[\sum_t \gamma^{t-1}r_t]$, where $\gamma \in [0,1]$ denotes the discount factor. 
The Value function $V_{\pi}:s \mapsto \mathbb{R}$ denotes estimated cumulative rewards, starting at state $s$ and following policy $\pi$. 
In deep reinforcement learning, policy $\pi$ and value function $V$ is handled by using a (deep) neural network as a function approximator, because the state and action space, $S, A$, are prohibitively large in most interesting tasks. 

Suppose a neural network parameterized by $\theta$ takes state $s$ as its input and yields a probability distribution on actions $\pi(s)$ as well as an estimate of value function $v(s)$ as its output.  Given a set of state transitions $\langle s_t,a_t,r_t,s_{t+1}\rangle$, (one-step) \emph{Advantage Actor Critic} updates $\theta$ for such direction that increases the probability of a good action and moves $v(s_t)$ closer to $r_t + \gamma v(s_{t+1})$:
\begin{align*}
\nabla_\theta J_\pi(\theta) &= \nabla_\theta \ln(\pi(a_t|s_t;\theta)) A(s_t,a_t;\theta),\\
\nabla_\theta J_v(\theta) &=  - \nabla_\theta v(s_t;\theta) \left(r_t + \gamma v(s_{t+1}) - v(s_t) \right) 
\end{align*}
where $A(s_t,a_t)$ is advantage of taking action $a_t$ at state $s_t$, and $A(s_t,a_t) = Q(s_t,a_t) - V(s_t) \approx r_t + \gamma v(s_{t+1}) - v(s_t)$. 
To prevent premature convergence, the entropy of the policy is often added to the objective function~\cite{williams1992simple, MnihBadiaMirzaGravesLillicrapHarleySilverKavukcuoglu2016}. 

\subsubsection{Application to Two-Player Games}
In typical application of RL to two-player board games, the ``agent'' stands for the player who is learning, and the ``environment'' includes both the opponent, and the rules of a game. 
The reward is given only at the termination of a game, as 1, 0, -1 for win, draw, loss, respectively.  
Given that agents are not enhanced by game-specific knowledge, the agent as well as its opponent must start as random players. 
AlphaZero~\cite{AlphaZero} begins by gathering game records of random players, then gradually updates the agent by their experiences and periodically replace the opponent by the learn agent.
Although changing the opponent along during learning makes the environment non-stationary and may introduce difficulty in training, it is effective to explore the challenging part of the state space and to improve the agent's strength. 
In our work, we applied reinforcement learning to two-player games in a similar way as AlphaZero. 

There are several major achievements in imperfect information games, including Texas Hold'em, Marjong, and StarCraft II.
However, their playing strength is supported by human game records in the target domain~\cite{VinyalsBabuschkinCzarnecki2019,Suphx}, or by methods based on counterfactual regret minimization~\cite{BowlingBurchJohansonTammelin2015} that is usually not applicable to games due to an intractable number of information sets growing almost exponentially along with the length of a game history.

\subsubsection{Residual Convolutional Neural Network}
A Convolutional Neural Network (CNN) is standard technique to handle images.  
It is also used for making RL agent in video games to understand the game screen such as Atari~\cite{MnihKavukcuogluSilverGravesAntonoglouWierstraRiedmiller2013}. 
Residual Neural Networks~\cite{ResNet} -- or ResNet -- is an enhancement for CNN to make learning efficient by adding residual path between layers.
AlphaZero incorporated ResNet for RL agents in Go, Chess, or Shogi.
We introduced alternative network for Catan and use ResNet as a baseline in comparison.

\subsection{Rules of Two-Player Catan}

The rules of Catan used for our research, as well as the naming conventions, matches the one of Catan Studio, Inc and Catan GmbH's official 5th edition rules\footnote{https://www.catan.com/service/game-rules}, but with only two players and no trading between them.
In this game, both players compete to colonize an island represented by a board of hexagonal tiles. There are 5 resource types -- Brick, Lumber, Ore, Grain, and Wool --  which can be spent to make various actions.
The first player to reach 10 Victory Points (VP) or more is considered the winner. VP can be acquired by various means: placing settlements (1VP) or cities (2VP) on the board, having the longest road or largest army (2VP), or special development cards (1VP).

The island of Catan is represented as a board of 19 \textit{land} hexagonal tiles called \textit{hexes}, randomly placed when setting up the game. Tiles can either represent a desert, or produce one of the 5 resources, in which case they will be assigned a number between 2 and 12.
We will call the edge of a hex a \textit{path}, and its corner an \textit{intersection}.

At the beginning of the game, each player places 2 settlements, each with an adjacent road, in the following order: player A, player B, player B, player A. Settlements must be placed on intersections and can not be next to one another.
%
\begin{table}[ht]\small
  \centering
  \caption{Actions in Catan. The first column denote the type of each action: `dice' means it is mandatory and once, `+' can be performed in any order and any number of times after `dice', `*' indicates it is allowed only once, but at any time}
  \label{tab:actions-in-turn}
  \smallvsqueaze
  \begin{tabular}{c|p{.93\linewidth}}\hline
    type & effect\\\hline
dice  & \emph{Roll} two 6-sided dice. If the sum is 7, every player with 7 or more resources must discard half of them, and the current player moves the robber.
    Otherwise, every hex with the corresponding sum will produces resources, giving one resources to each settlement adjacent to it, and two for cities.\\
+   & Buy a \emph{Road}. Spend Brick + Lumber to place one on a path, next to another road.\\
+   & Buy a \emph{Settlement}. Spend Brick + Lumber + Grain + Wool to place a settlement next to a road, on an intersection surrounded by unoccupied intersections.\\
+   & Buy a \emph{City}. Spend 3 Ores + 2 Grains to improve an already placed settlement into a city.\\
+   & Buy a \emph{Development Card}. Spend Ore + Grain + Wool to draw one card from the development pile, look at it, and add it to your hand at the end of your turn\\
+   & \emph{Trade} resources with the bank. The default ratio is four of the same resource for any one resource, but having a settlement or city on a harbor can reduce the rate to 3:1 or 2:1.\\
*   & Use a \emph{Development Card}. The card is revealed and consumed (see Table~\ref{tab:dev-cards}).\\\hline
  \end{tabular}
\end{table}
During each turn, a player can take a sequence of actions under constraints, listed in Table~\ref{tab:actions-in-turn}.
The robber is a piece located on a hex that prevents production on it. After rolling a 7 or using a Knight development card, the current player must move the robber to a new hex. If the other player has a settlement or city adjacent to this new location, the current player forcibly takes a random resource.
Development cards are shuffled into a face down pile at the beginning of the game. Each has one of the effects listed in Table~\ref{tab:dev-cards}. 
\begin{table}[ht]\small
  \centering
  \caption{Development cards}
  \label{tab:dev-cards}
  \smallvsqueaze
  \begin{tabular}{r|p{.8\linewidth}}\hline
    Knight card& Move the robber (see \textit{the robber}), and increment army size\\
    Road building& Place two roads for free\\
    Year of Plenty& Take two resources from the bank\\
    Monopoly& The opponent gives you all their resources of a stated type\\
    Victory Point& Get one victory point\\\hline
  \end{tabular}
\end{table}
There is no simple winning strategy. Basically, a stable and varied production of resources is beneficial to obtain VP. Thus, players should prioritize placing their settlements in intersections surrounded by balanced resources and high production chance (with numbers around 7), near promising un-exploited areas or on interesting harbors. However, what strategy is good highly depends on the board configuration and random dice rolls.

\subsection{JSettlers and Research on Catan}
JSettlers~\cite{JSettlers2} is an open-source Java implementation of the Catan rules.
Among the many features the environment offers, it contains a hand-coded heuristic-based agent very often used as a base-line in Catan research.
In this study, we used version 2.2.00 (released on the 3rd of March 2020), and kept the default agent type proportions: 30\% of ``smart-bots'' and 70\% of ``fast-bots''. In the rest of the paper, we will call this agent \textit{jsettler}.
We used JSettlers only for evaluation purpose (not in training) due to its slow execution speed.
Note that its rules do not perfectly match the official rules (e.g. it doesn't include the 19 resources limit), but ``official'' agents can play in JSettlers with minor adjustments.

The earliest agent used Model Trees trained through self-play~\cite{RLstrats_ModelTree}. It hasn't been compared to JSettlers, but against a human, the author of the paper.

Szite I. et al. used Monte-Carlo Tree Search in a perfect-information variation of the game~\cite{MCTS}. Their agent manages to obtain a 27\% winrate with 1000 simulations, and 49\% winrate with 10000 simulations, when playing against 3 \textit{jsettlers}. However, their method cannot be applicable in the original, i.e., imperfect information, rule.

We have found two papers that used Deep Reinforcement Learning, but they focused only on a subset of actions: trading. They both used a \textit{jsettler} agent as a base and replaced its trading behavior, and compared its performance against 3 \textit{jsettlers}: one achieved 49\% winrate with Deep Q-Learning~\cite{DialogueDRL}, the other 52\% winrate with online Deep Q-Learning with LSTM~\cite{DRRL}.

In this paper, our agents do not learn trading -- refusing all offers and never
initiating negotiation -- due to the limitation in our computational resources. We
assert it is still fair as it does not introduce any advantage for our agents. We also limit the number of players to two instead of three or more.
We argue that the task is still challenging, and to our best knowledge, this is
the first study in which agents trained by reinforcement learning
without domain knowledge successfully outperform \textit{jsettlers}.

\section{Our approach}

\subsection{Training process}

\subsubsection{Modified Advantage Actor Critic}

Our agent is mostly based on Advantage Actor Critic. However, to speed up the learning and diversify experiments, some parallelism has been added. (Although there are similarities, it isn't A3C~\cite{MnihBadiaMirzaGravesLillicrapHarleySilverKavukcuoglu2016}.)

Instead of playing one game on a single thread, experiences are acquired by 16 parallel workers, each playing 8 games at the same time. Each worker will cycle through its games, playing one move and saving the experience. Once a batch of 64 moves has been generated, the worker sends it to the trainer.

Simultaneously, another process is training the neural network on the batches it receives. After each update, the trainer propagates the weights to all workers.

Since a batch is not sent until it is full, some of the earliest experiences it contains were played with a slightly older policy. However, since this only represents a fraction of the batch, and the tardiness is of only a dozen of training steps, the off-policy aspect can be considered negligible.

\subsubsection{Self-play against past versions}

Our agent is trained against a past version of itself, but each worker uses a different time stamp.
Every 50 training steps ($50 \times 1000 \times 64$ moves, around an hour), the worker with the oldest opponent will update its policy to the most recent one.

This has many different advantages, and its efficiency has been shown in Figure~\ref{fig:selfplay}. This way, the opponents:
\begin{itemize}
    \item change ``slowly'': every 50 steps, only one agent among 16 is changed
    \item are varied: they have the behaviors the past trained agent had spanning over 750 steps (${\sim}18$ hours)
    \item match the level of the trained agent: the newest opponents are at a level very close to that of the trained agent
\end{itemize}

\subsubsection{Policy activity loss}

In order to encourage exploration, we can add an entropy gradient as mentioned in 3.2. However entropy only affects legal actions, as the others being masked. In Catan, some actions are very rare and might be playable only once every couple of games (e.g. Monopoly). In order to prevent these actions' probabilities from drifting into near-zero during the many weights updates, we added a L2 activity loss on the policy layer. This loss is applied directly on the logits $\left\{p_i\right\}_{i}$, the raw output before a softmax activations maps them to probabilities.
%
%
Thus, it will control the policy by pulling the average towards 0 and curbing absurdly high or low probabilities. Its empirical effect on the stability of the learning is shown in Figure~\ref{fig:activation}.

This final gradient is (with empirical hyper-parameters defined in Table~\ref{tab:hyperparams}):
\[
\nabla_\theta J = \alpha_\pi \nabla_\theta J_\pi(\theta)
+ \alpha_v \nabla_\theta J_v(\theta)
+ \alpha_H \nabla_\theta \sum_{a} \Tilde{\pi}(a|s) \ln{\Tilde{\pi}(a|s)}
+ \alpha_p \nabla_\theta \sum_i p_i^2
\]

\subsection{Encoding and network structure}

\subsubsection{Adapted Convolutional Neural Network for Hexagonal board}

A regular board of catan contains 19 hexes, 72 paths and 54 intersections. This number being large for fully-connected networks, we would like to take advantage of the regularity of the board by using Convolutional Neural Network (CNN) layer. However, typical CNN are tailored specifically for grid-like layouts. There exist some tricks to fit a hexagonal grid into a regular grid, but the existing ones are not directly applicable in Catan as we also need to represent the paths (edges) and intersections (vertices).

\begin{figure}[ht]
\centering
\begin{minipage}[b]{0.4\linewidth}
\begin{center}
\includegraphics[width=\textwidth]{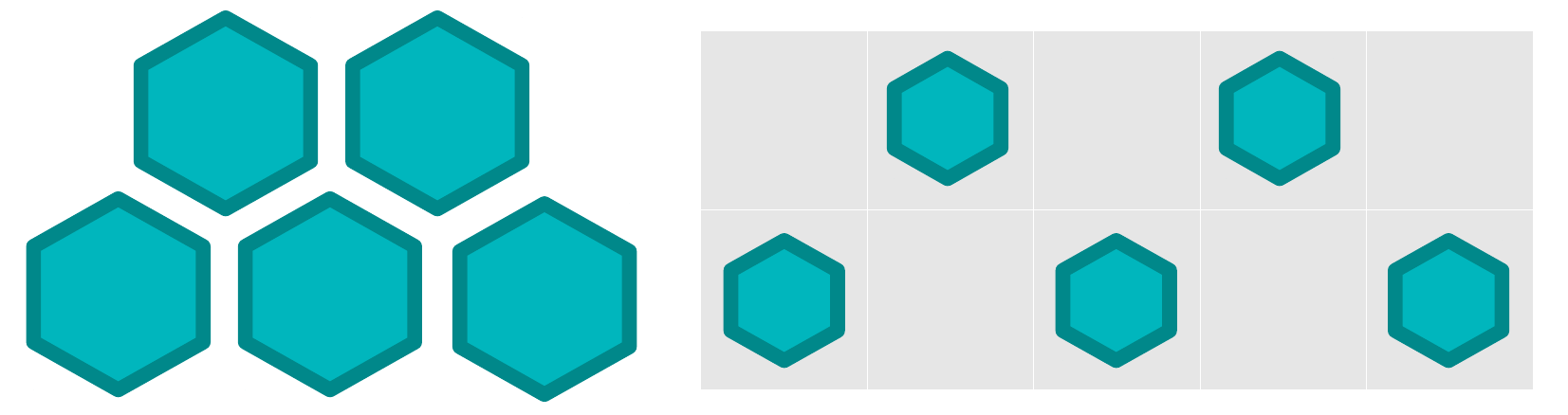}
\bigvsqueaze
\caption{Double coordinate} \label{fig:double}
\end{center}
\end{minipage}
\hspace{8mm}
\begin{minipage}[b]{0.4\linewidth}
\begin{center}
\includegraphics[width=\textwidth]{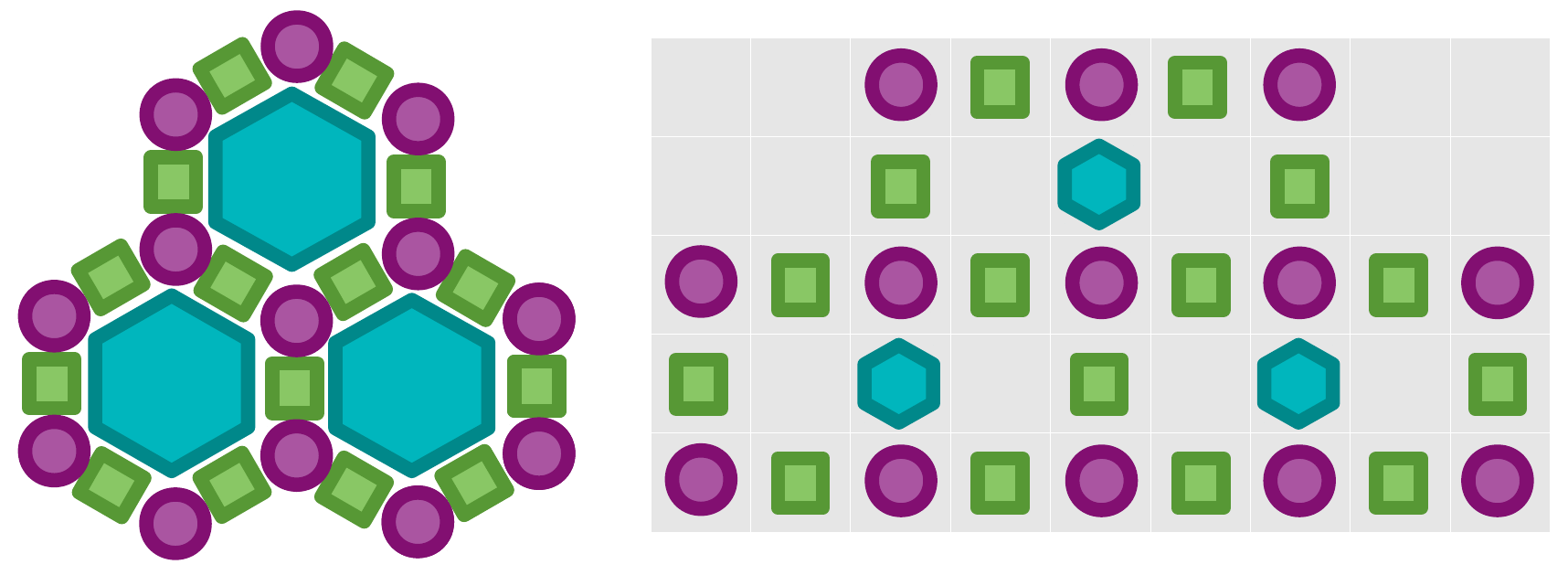}
\bigvsqueaze
\caption{Brick coordinate} \label{fig:brick}
\end{center}
\end{minipage}
\par
\end{figure}

\begin{figure}[ht]
\centering
\includegraphics[width=0.24\textwidth]{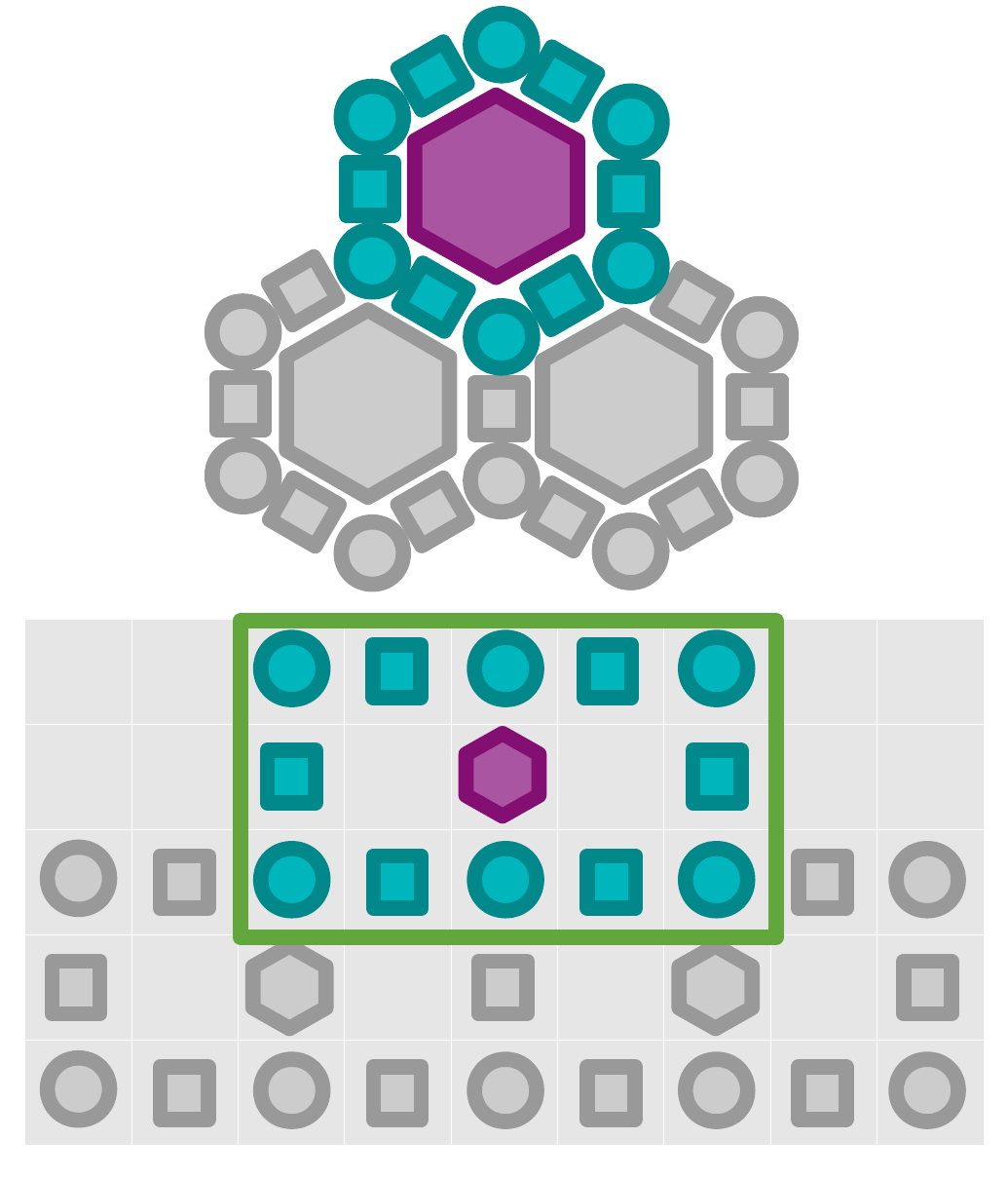}
\hspace{3mm}
\includegraphics[width=0.24\textwidth]{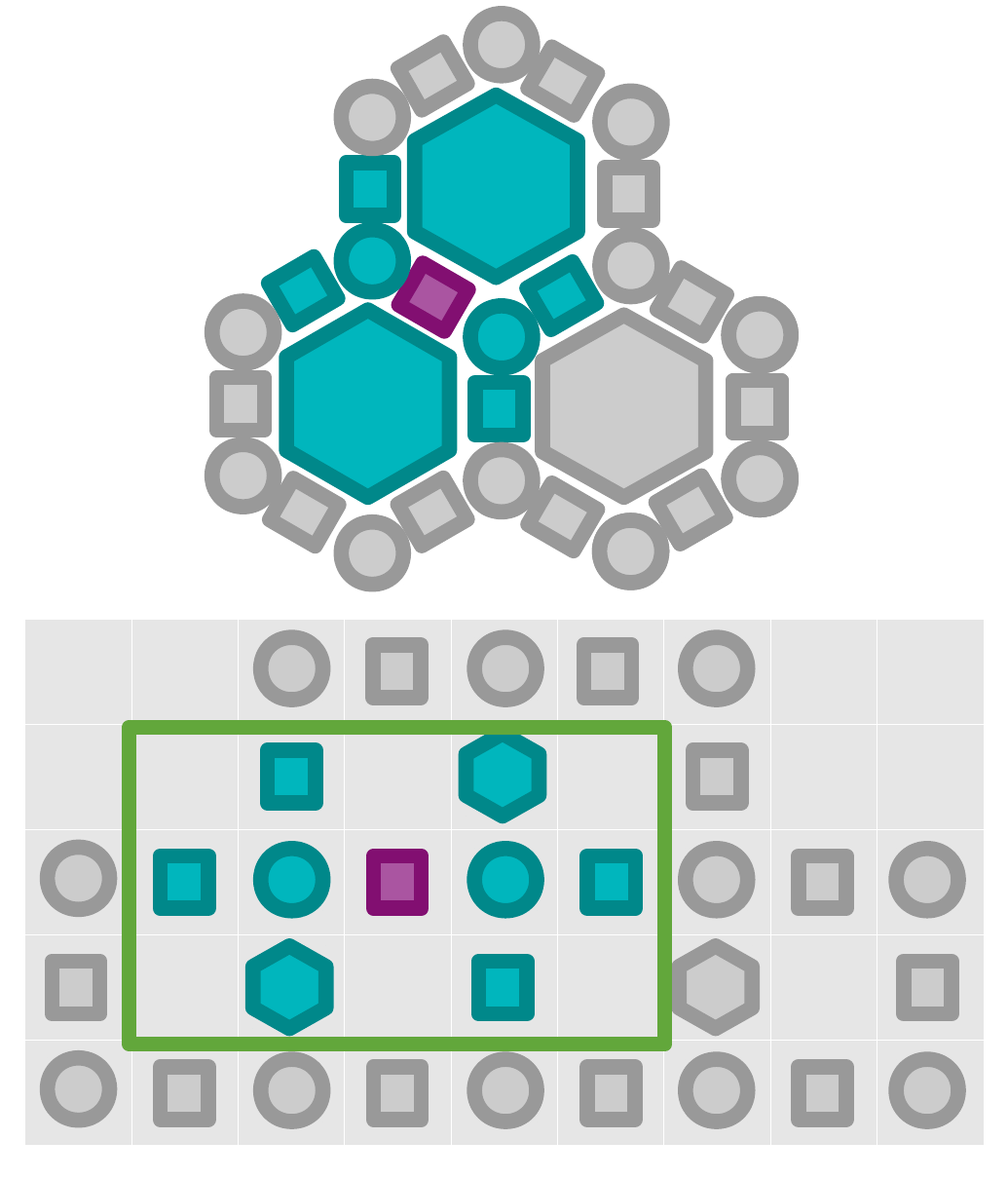}
\hspace{3mm}
\includegraphics[width=0.24\textwidth]{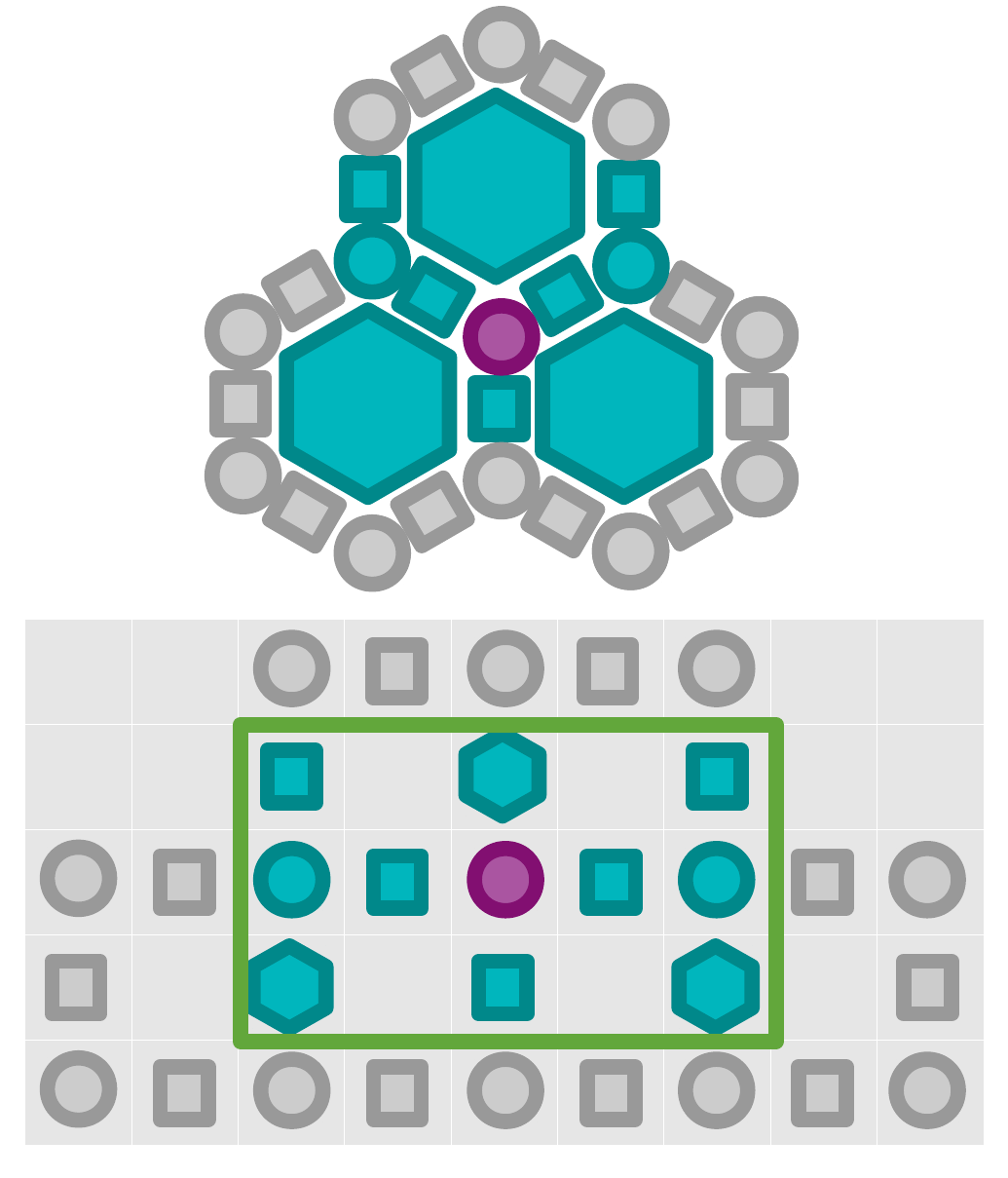}
\vsqueaze
\caption{$5 \times 3$ kernel on brick coordinate} \label{fig:brickSelect}
\par
\end{figure}

Our idea, that we called ``brick coordinate'' (Fig.~\ref{fig:brick}) was inspired by the double coordinate method (Fig.~\ref{fig:double}). By using a $5 \times 3$ kernel, the neighbors in brick coordinate considered by the CNN are very similar to the actual neighbours on the hexagonal board (Fig.~\ref{fig:brickSelect}).

Furthermore, unlike board games like chess or go, where every position represents the same type of cell, Catan has hexes, paths and intersections that have radically different behaviors, neighbors, and features. To prevent the convolution from processing them equivalently, we separate features or actions of different types in different channels.

\subsubsection{Cross dimensional neural network}

In most games where CNN can be used to efficiently process the input state, non spacial features that don't correspond to any position can be added as extra channels (e.g. turn channels in AlphaZero).
However, Catan has a lot of such features, as well as actions that are completely unrelated to a position on the board (e.g. playing a development card, trading, or ending one's turn).

Intuitively, we would want to handle them using fully connected layers, but doing two networks in parallel degenerates perfomance.
To overcome this problem, we propose using \textit{Cross Dimensional Neural Network}. The idea is to combine two networks in parallel, each tailored for processing neurons of different dimensions, and inter-connect them to propagate information from one type into the other.

\begin{figure}[ht]
\centering
\begin{minipage}[b]{0.53\linewidth}
\begin{center}
\includegraphics[width=\textwidth]{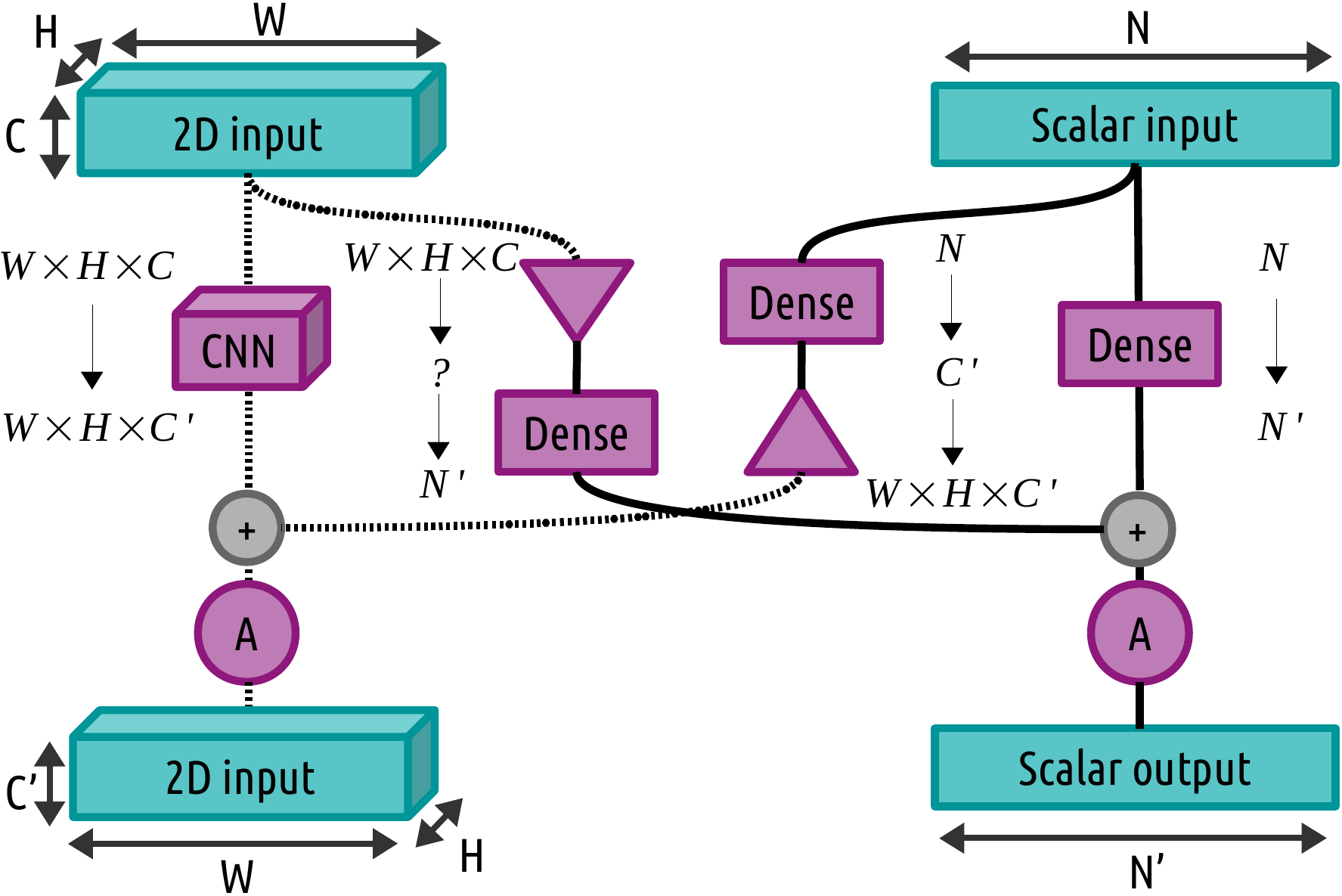}
\bigvsqueaze
\caption{Base Cross Dimensional (Xdim) layer used in our experiment: $\bigtriangledown$ is deflation, $\bigtriangleup$ is inflation, $A$ is activation} \label{fig:xdimDetails}
\end{center}
\end{minipage}
\hspace{5mm}
\begin{minipage}[b]{0.33\linewidth}
\begin{center}
\includegraphics[width=\textwidth]{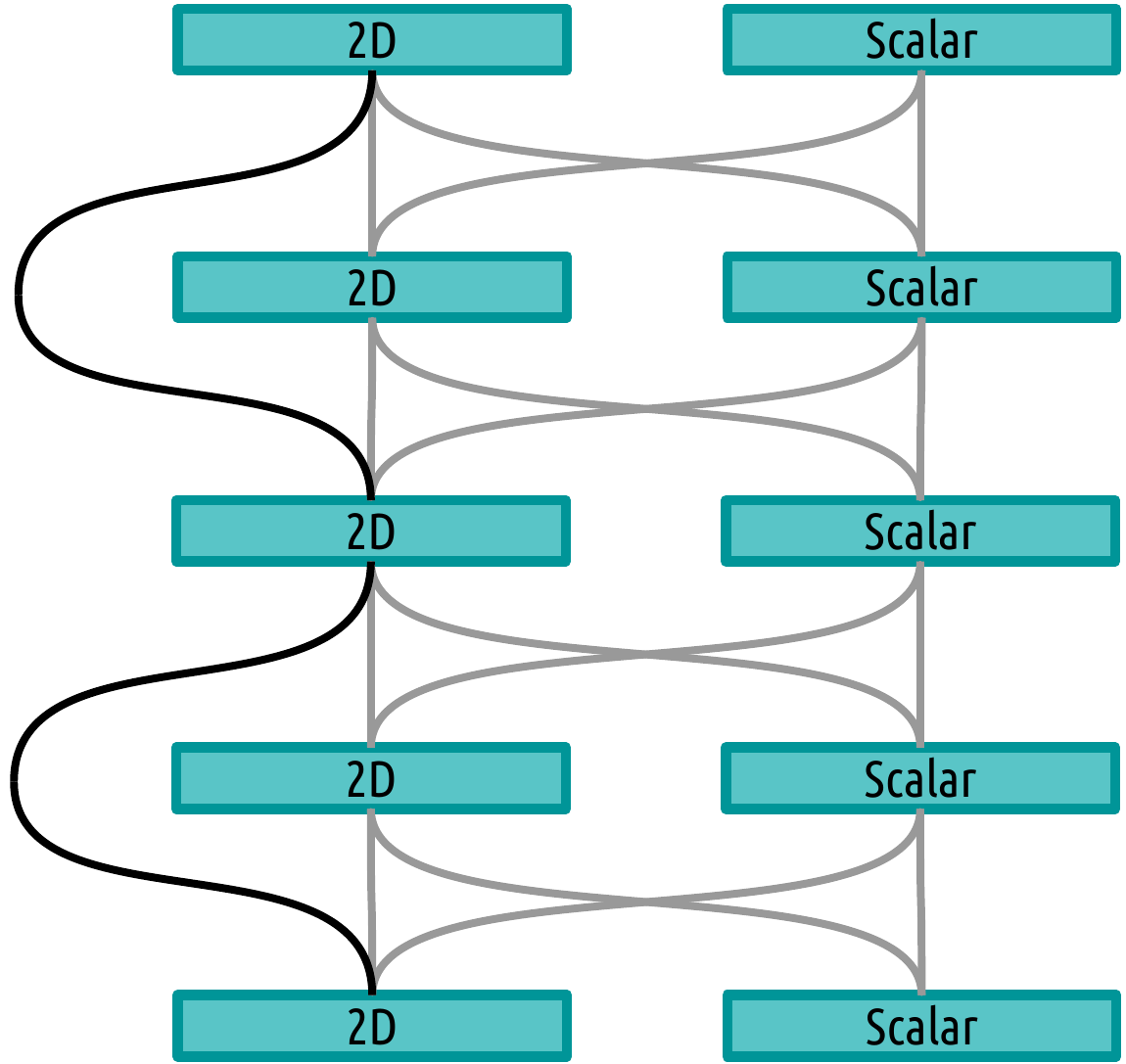}%
\vspace{1mm}%
\caption{Residual Cross Dimensional layer model} \label{fig:xdimresModel}
\vspace{3.7mm}
\end{center}
\end{minipage}
\par
\end{figure}

For example, in Catan, we would have one series of layers for the 2-dimensional features, one for scalar features, and interconnections between them (Fig.~\ref{fig:xdimDetails}).

In order to connect features of different dimensions (i.e. brick coordinate channels and non-spacial features), we will need to either ``inflate'' or ``deflate'' them, and adjust the shape with dense layers. In this paper, we used the following:
\begin{itemize}[topsep=0.5mm]
    \item For inflation, each scalar value is converted to a channel filled with that value
    \item For deflation, each channel is reduced to two scalars: its average and variance
\end{itemize}
Then, we can get the values for each head of the Cross dimensional neural network by summing the output of both sources and applying an activation.

As shown in Figure~\ref{fig:xdimresModel}, it is also possible to incorporate Residual paths into Cross Dimensional layers. Note that the sum is made before the activation.

\subsubsection{Encoding of features and actions}

A summary of how the state space and action space can be seen on Table~\ref{tab:state} and Table~\ref{tab:action} respectively.

\begin{table}[!ht]
    \newcommand{\subrow}[1]{\hspace{0.25cm}\footnotesize\textit{#1}}
    \newcommand{\subnum}[1]{\footnotesize\textit{#1}}
    \newlength{\lsq}
    \setlength{\lsq}{-0.65mm}
    \begin{minipage}{.51\linewidth}
        \caption{Input - Observable State} \label{tab:state}
        \smallvsqueaze
        \centering
        \begin{tabular}{lr}
        \hline
        \textbf{Board (num. of 2D channels)}          & \textbf{17}  \\ \hline
        Hexes                                         & 7            \\[\lsq]
        \subrow{Is Desert}                            & \subnum{1}   \\[\lsq]
        \subrow{Production for each resource}         & \subnum{5}   \\[\lsq]
        \subrow{Thief}                                & \subnum{1}   \\ \hline
        Paths                                         & 2            \\[\lsq] 
        \subrow{Road for each player}                 & \subnum{2}   \\ \hline
        Intersections                                 & 8            \\[\lsq]
        \subrow{Harbors}                              & \subnum{6}   \\[\lsq]
        \subrow{Settlement or city}                   & \subnum{2}   \\ \hline \hline
        \textbf{Others (dim. of vector)}              & \textbf{45}  \\ \hline
        Self                                          & 27           \\[\lsq]
        \subrow{Resources}                            & \subnum{5}   \\[\lsq]
        \subrow{Pieces left}                          & \subnum{3}   \\[\lsq]
        \subrow{Army size}                            & \subnum{1}   \\[\lsq]
        \subrow{Held development cards (new + old)}   & \subnum{10}  \\[\lsq]
        \subrow{Access to each harbor}                & \subnum{6}   \\[\lsq]
        \subrow{Largest Army and Longest Road}        & \subnum{2}   \\ \hline
        Opponent                                      & 8            \\[\lsq]
        \subrow{Resource and Development card total}  & \subnum{2}   \\[\lsq]
        \subrow{Pieces left}                          & \subnum{3}   \\[\lsq]
        \subrow{Army size}                            & \subnum{1}   \\[\lsq]
        \subrow{Largest Army and Longest Road}        & \subnum{2}   \\ \hline
        General                                       & 6            \\[\lsq]
        \subrow{Bank resources}                       & \subnum{5}   \\[\lsq]
        \subrow{Development Card Pile}                & \subnum{1}   \\ \hline
        Phase                                         & 4            \\[\lsq]
        \subrow{Has Rolled}                           & \subnum{1}   \\[\lsq]
        \subrow{Has development card been played}     & \subnum{1}   \\[\lsq]
        \subrow{Using RoadBuilding or YearOfPlenty}   & \subnum{2}   \\ \hline
        
        \end{tabular}
    \end{minipage}%
    \hspace{1mm}
    \begin{minipage}{.47\linewidth}
      \centering
        \caption{Output - Prob of Actions} \label{tab:action}
        \smallvsqueaze
        \begin{tabular}{lr}
        \hline
        \textbf{Board (num. of 2D channels)}          & \textbf{5}  \\ \hline
        Hexes                                         & 2            \\
        \subrow{Move thief and steal}                 & \subnum{1}   \\
        \subrow{Move thief without stealing}          & \subnum{1}   \\ \hline
        Paths                                         & 1            \\ 
        \subrow{(Buy and) Place road}                 & \subnum{1}   \\ \hline
        Intersections                                 & 2            \\
        \subrow{(Buy and) Place Settlement}           & \subnum{1}   \\
        \subrow{Buy and Place City}                   & \subnum{1}   \\ \hline \hline
        \textbf{Others (dim. of vector)}              & \textbf{117} \\ \hline
        Phase                                         & 2            \\
        \subrow{Roll dice}                            & \subnum{1}   \\
        \subrow{End turn}                             & \subnum{1}   \\ \hline
        Resources                                     & 90           \\
        \subrow{Discard (4 cards to keep)}            & \subnum{70}  \\
        \subrow{Bank trade}                           & \subnum{20}  \\ \hline
        Development Card                              & 22           \\
        \subrow{Buy development card}                 & \subnum{5}   \\
        \subrow{Activate Knight}                      & \subnum{5}   \\
        \subrow{Activate Road Building}               & \subnum{1}   \\
        \subrow{Activate Year of Plenty}              & \subnum{1}   \\
        \subrow{Choose free resource}                 & \subnum{5}   \\
        \subrow{Play Monopoly (each resource)} & \subnum{5}   \\ \hline
        \end{tabular}
    \end{minipage} 
\end{table}

We can see that Catan needs a much more complex representation for state and actions than those of chess and shogi used in AlphaZero. For discards (after a roll of 7), we introduced a ``keep 4 resources'' abstraction, using only 70 representative actions, rather than the 1\,599\,979 discarded resources actions (the cardinal of $\left\{b,l,o,g,w \in \left\{0,1 \ldots 19\right\}^5 \mid 3 \leq b+l+o+g+w \leq 47 \right\}$, where each of $b,l,o,g,w$ stands for a resource type). The 70 actions perfectly covers usual situations, where only four resources are kept. Even in exceptional cases (when holding more than 8 resources), the agent behaves robustly by randomly picking additional resources, after having saved the best four.




\section{Experiments and Results}

For the experiments, 3 different types of neural network architecture where used, each with 6, 8, and 10 layers (alternating tanh and leaky-ReLU activations):
\begin{itemize}[noitemsep, topsep=0.5mm]
    \item \textbf{CNNRes}, baseline, a 40-channel CNN with ResNet (without Xdim)
    \item \textbf{Xdim}, our method, using 15 2D-channels and 40 non-spacial neurons \linebreak ($C = 15$ and $N = 40$ on Figure~\ref{fig:xdimDetails})
    \item \textbf{XdimRes}, variation of our method with Residual paths
\end{itemize}

The hyper parameters used are described in Table~\ref{tab:hyperparams}. The reward is given once a game is finished, and is +0.75 for winning (resp. -0.75 for loosing) and +0.02 for every VP over the opponent's (resp. -0.02 for every VP behind).
\begin{table}[!ht]
    \centering
    \caption{Hyper Parameters}
    \label{tab:hyperparams}
    \vsqueaze
    \renewcommand{\arraystretch}{0.9}
    \begin{minipage}[t]{.5\linewidth}
        \centering
        \begin{tabular}[t]{p{.7\linewidth}|>{\hspace{1mm}}p{.225\linewidth}}
        \multicolumn{2}{c}{Learning rate}\\
        \hline
        Initial value & $3 \times 10^{-3}$  \\
        Inverse decay / training step & $2 \times 10^{-3}$  \\
        \hline
        \end{tabular}
        
        \vspace{2mm}
        \begin{tabular}[t]{p{.7\linewidth}|>{\hspace{1mm}}p{.225\linewidth}}
        \multicolumn{2}{c}{Reward}\\
        \hline
        Winning reward & $\pm0.75$  \\
        VP difference reward & $\pm0.02$  \\
        \hline
        \end{tabular}
    \end{minipage}%
    \hspace{1mm}
    \begin{minipage}[t]{.47\linewidth}
        \centering
        \begin{tabular}[t]{p{.5\linewidth}p{.2\linewidth}|>{\hspace{1mm}}p{.225\linewidth}}
        \multicolumn{3}{c}{Gradient Factor}\\ 
        \hline
        Policy & $\alpha_\pi$ & $1 \times 10^{0}$  \\
        Value function & $\alpha_v$ & $1 \times 10^{3}$  \\
        Entropy & $\alpha_H$ & $1 \times 10^{-4}$  \\
        Policy activity loss & $\alpha_p$ & $1 \times 10^{-8}$  \\
        Weight L2-regu & $\alpha_\theta$ & $1 \times 10^{-4}$  \\
        \hline
        \end{tabular}
    \end{minipage}
    \renewcommand{\arraystretch}{1}
\end{table}

We used Tensorflow 2.1 compiled for CUDA 10.2, and the code was run on a 32 Core CPU\footnote{AMD Ryzen Threadripper 2990WX} with two GeForce GTX1080Ti 11GB GPU.
In order to generate experiences quickly, we implemented a minimal environment of Catan in the Rust language, focusing on execution speed.
To use it seamlessly with Tensorflow, we also turned it into a Python module using the PyO3 bindings.
It is open source and can be found at \emph{https://github.com/Swynfel/rust-catan}.

On the following figures, one \textit{training} corresponds to processing 1000 batches of 64 experiences each.

\subsection{Learning curves}

\begin{figure}[ht]
\centering
\begin{minipage}[b]{0.65\linewidth}
\includegraphics[width=\textwidth]{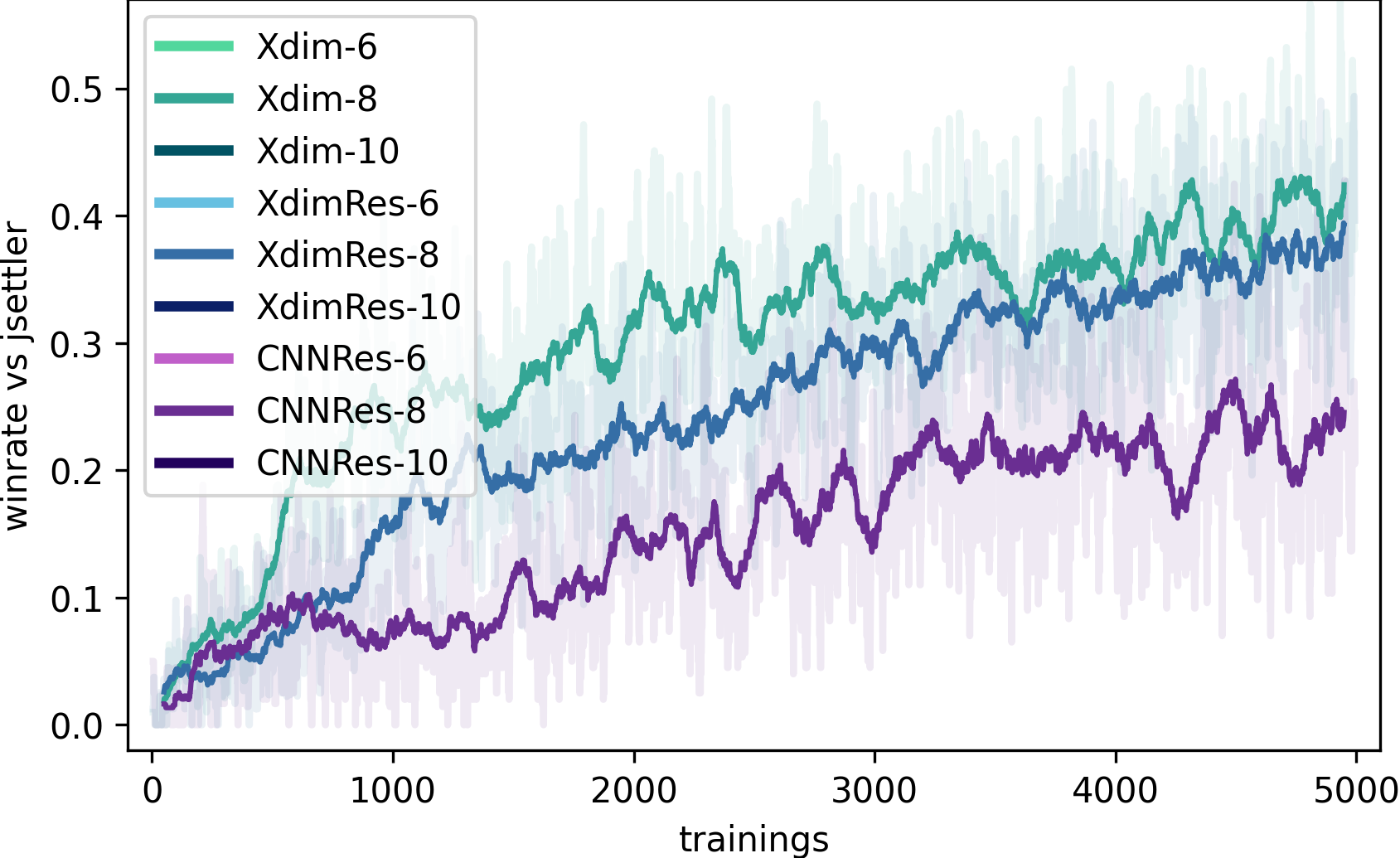}
\end{minipage}
\hspace{0mm}
\begin{minipage}[b]{0.3\linewidth}
\includegraphics[width=\textwidth]{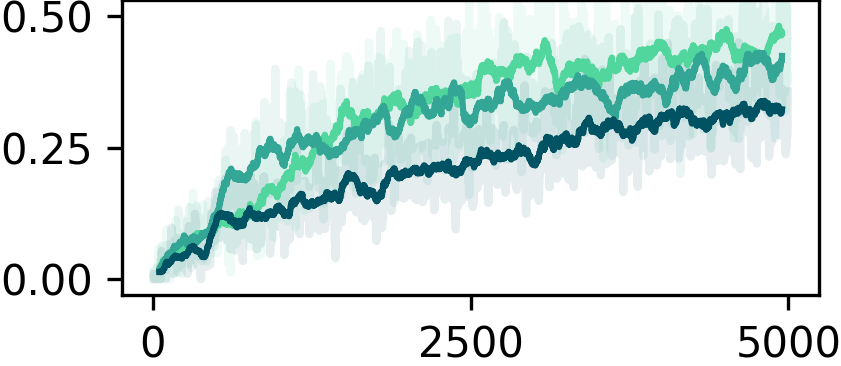}
\includegraphics[width=\textwidth]{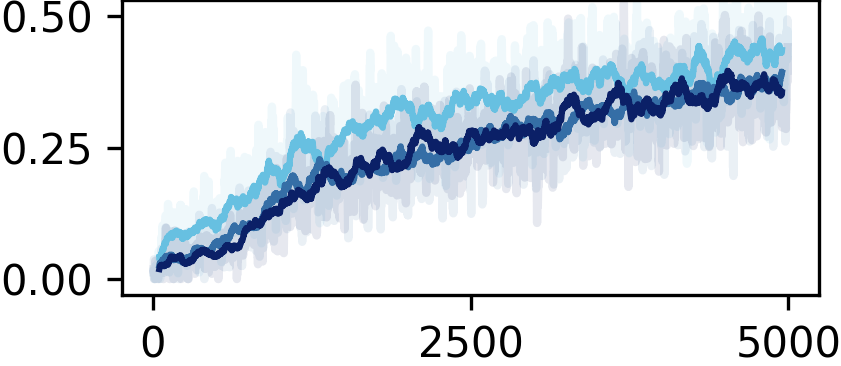}
\includegraphics[width=\textwidth]{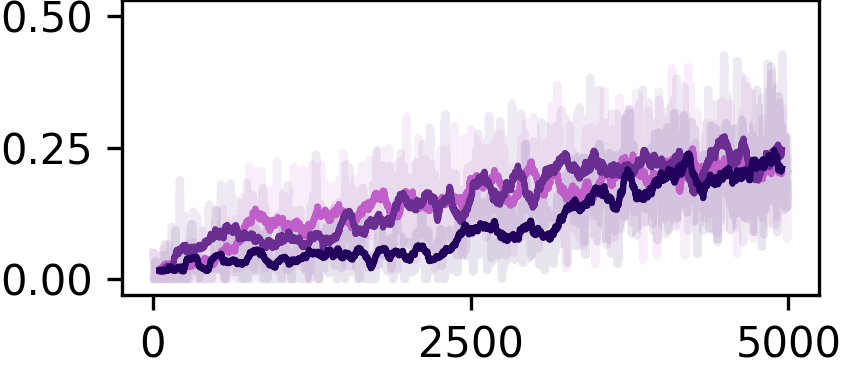}
\end{minipage}
\vsqueaze
\caption{Improvements of winrate against \textit{jsettler} during early training, comparisons of architectures (left) and comparisons of layer count for each architecture (right) } \label{fig:quick}
\par
\end{figure}

First, we looked at the learning curves of each model (Fig.~\ref{fig:quick}). %




Unsurprisingly, models with less layers learn faster, especially in the early steps of training.
When comparing architectures of different types, \textit{CNNRes} is lagging behind, but the other two models seems close. It is notable that even if ResNet are supposed to accelerate the early steps of training, we don't see such impact when used in conjunction with Xdim. It is even the opposite for models with a few layers.
Our hypothesis is that using Xdim already introduces a sort of shortcut (information can cross from 2D values to scalar, and back to 2D). This makes the ResNet not as useful, and its effect is negligible for networks that aren't very deep. However for models with 10 layers, we can see its impact again.

\subsection{Long term training results}

\begin{figure}[ht]
\centering
\includegraphics[width=0.48\textwidth]{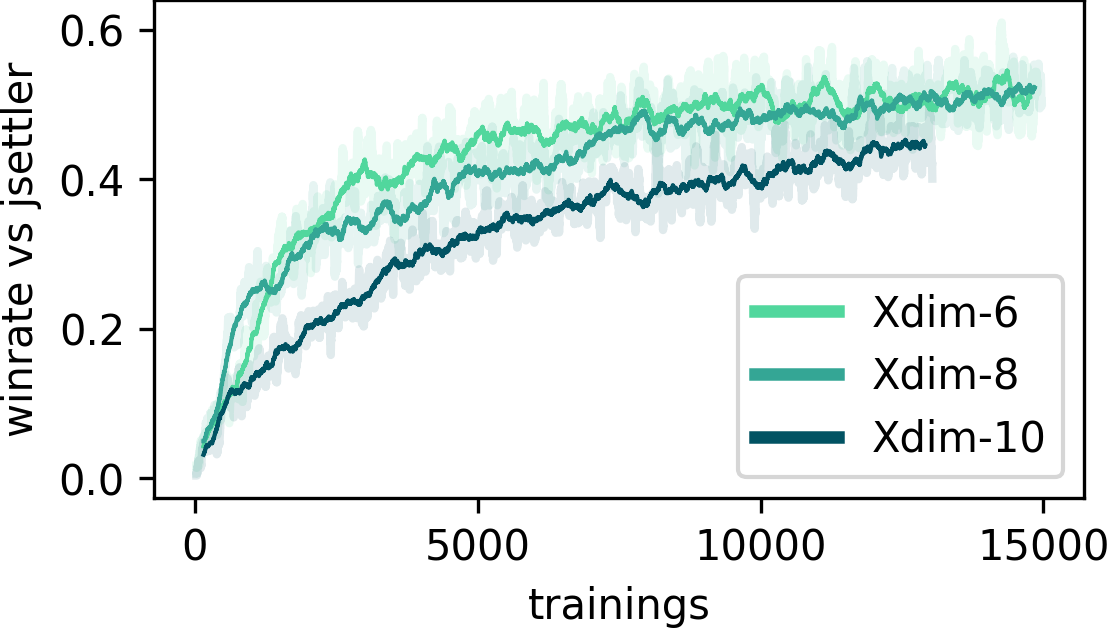}
\hspace{0mm}
\includegraphics[width=0.48\textwidth]{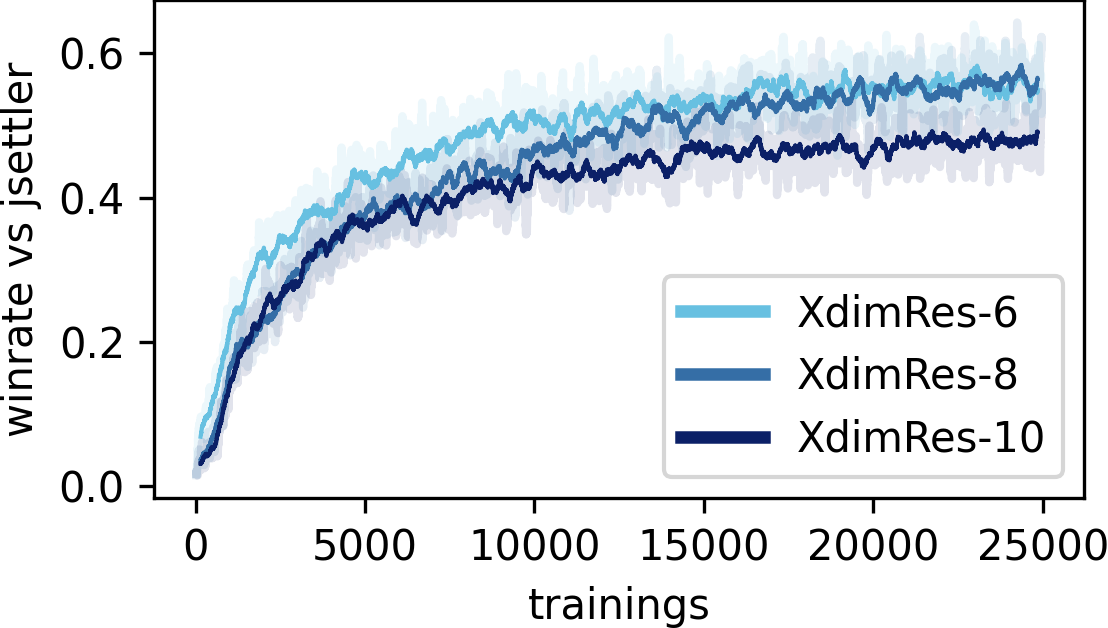}
\vsqueaze
\caption{Evolution of winrate against \textit{jsettler} during long training} \label{fig:long}
\par
\end{figure}
\begin{figure}[ht]
\centering
\includegraphics[width=0.48\textwidth]{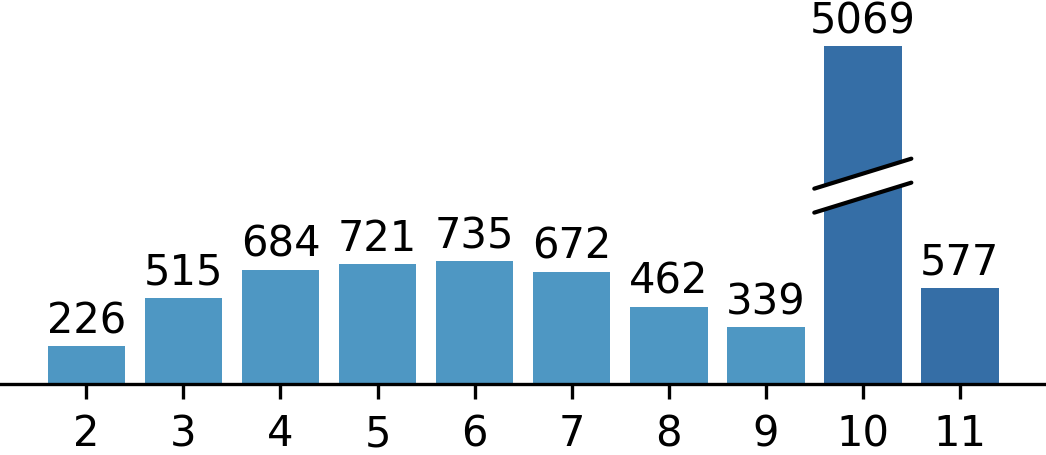}
\hspace{0mm}
\includegraphics[width=0.48\textwidth]{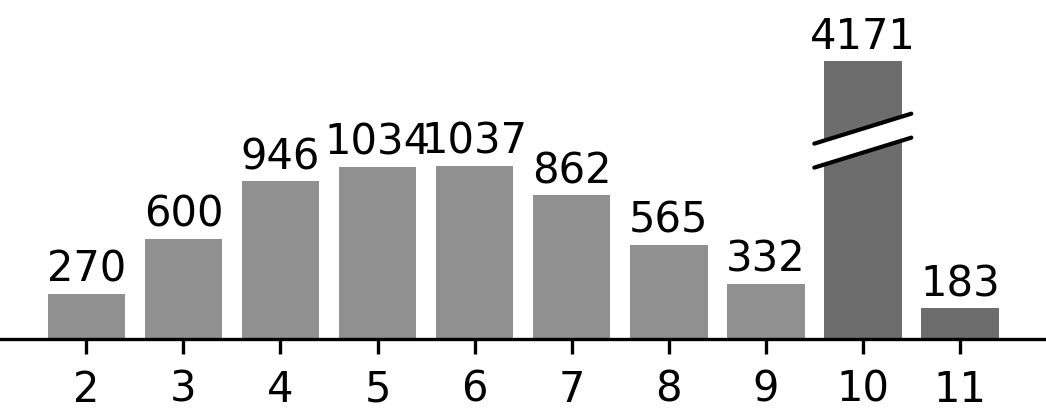}
\vsqueaze
\caption{Distribution of VP in 10\,000 games opposing XdimRes-8 after 30\,000 training steps (left) vs \textit{jsettler} (right)} \label{fig:histogram}
\par
\end{figure}
Since CNN layers are not promising even in the early stages of training, we only kept training the models using Cross Dimensional NN.
In the first 15000 training steps (around 3 weeks), we can see that the 8-layer models start catching up to the 6-layers one, and that using Residual Layers does help in the long run. We can also confirm that Xdim-6, Xdim-8, XdimRes-6, and XdimRes-8 all passed over 50\% win-rate (Fig.~\ref{fig:long}). When focusing on XdimRes-8 after 30000 training steps (approximately 5 weeks), we see even reached 56.5\% (Fig.~\ref{fig:histogram}). Thus we can confidently say our agent outperforms \textit{jsettler} in 1vs1.
%
%

\subsection{Ablation studies}

\begin{figure}[ht]
\centering
\begin{minipage}[t]{0.46\linewidth}
\begin{center}
\includegraphics[width=\textwidth]{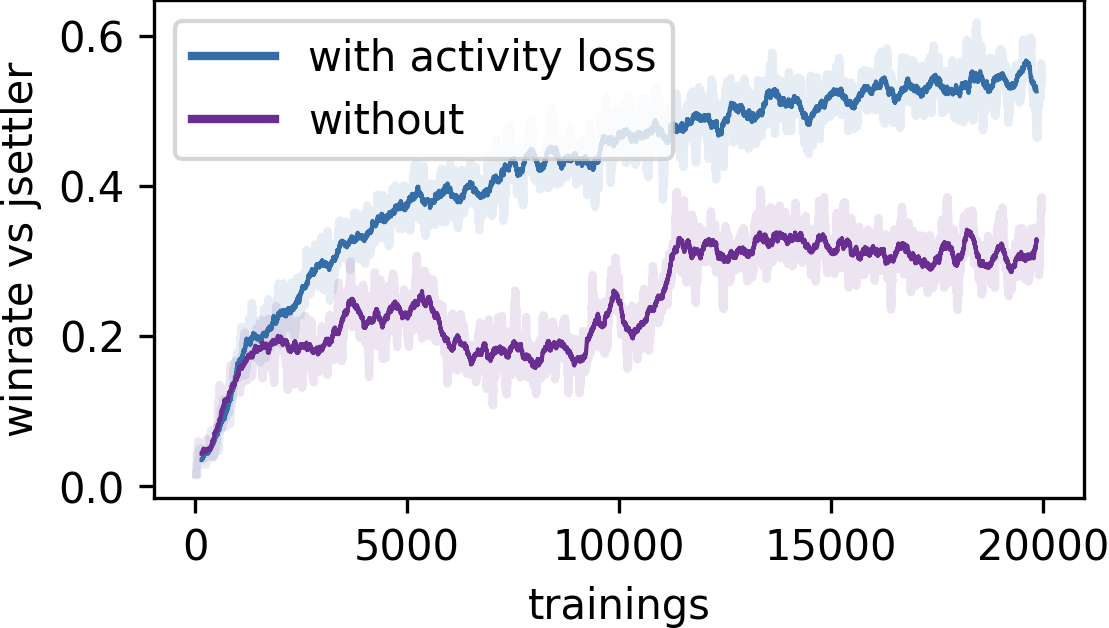}%
\vsqueaze
\caption{Learning curves of XdimRes-8 with and without policy activity loss}%
\label{fig:activation}%
\end{center}
\end{minipage}
\hspace{2mm}
\begin{minipage}[t]{0.46\linewidth}
\begin{center}
\includegraphics[width=\textwidth]{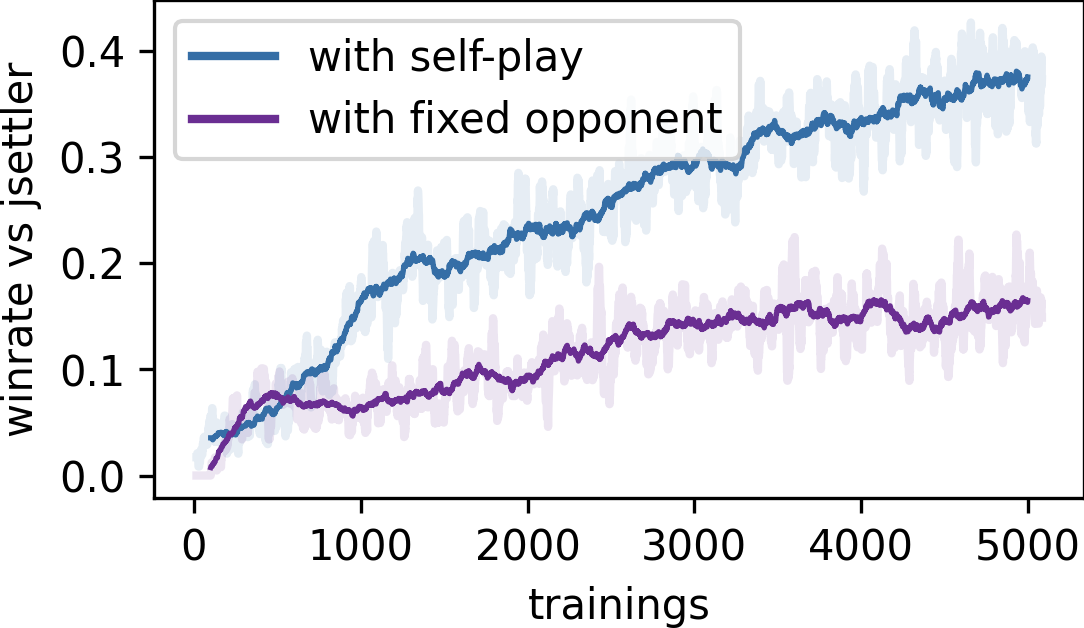}%
\vsqueaze
\caption{Learning curves with self-play (our method) and with fixed opponent}%
\label{fig:selfplay}%
\end{center}
\end{minipage}
\par
\end{figure}

We conducted two ablation studies. Figure~\ref{fig:activation} illustrates how removing the policy activity loss makes the training unstable. Figure~\ref{fig:selfplay} shows the importance of self-training against various opponents. The agent trained only against a fixed ``good'' opponent -- a copy of our agent after 10000 training steps -- has trouble learning at first as the opponent is too strong, only to overfit and play poorly against \textit{jsettler}, an unknown agent.

\section{Conclusion and Future works}

In this paper, we have shown how we can successfully overcome the difficulties of Catan:
The hexagonal board can be processed with CNN by using brick coordinate encoding; and the mix of positional and scalar features and actions can be handled with Cross Dimensional layers.
Combining these techniques, we created a Deep RL-based agent that reached 56.5\% win-rate against \textit{jsettler} with no prior target domain specific knowledge, trained only by self-play.

We would like to continue working on Catan but with the full rules: 4 players and allowing trades between players. These two aspects introduce interesting Multi-Agent-related challenges, such as prioritizing threats and collaborating actions (e.g. trading). Extending our agent to handle theses cases can offer lots of insights on how to build robust AIs that can interact with others in many ways.


\bibliographystyle{splncs04}
\bibliography{catan}
\end{document}